\documentclass[twocolumn]{article}
\usepackage{arxiv}

\usepackage[utf8]{inputenc} % allow utf-8 input
\usepackage[T1]{fontenc}    % use 8-bit T1 fonts
\usepackage{hyperref}       % hyperlinks
\usepackage{url}            % simple URL typesetting
\usepackage{booktabs}       % professional-quality tables
\usepackage{amsfonts}       % blackboard math symbols
\usepackage{nicefrac}       % compact symbols for 1/2, etc.
\usepackage{microtype}      % microtypography
\usepackage{cleveref}       % smart cross-referencing
\usepackage{lipsum}         % Can be removed after putting your text content
\usepackage{graphicx}
\usepackage[numbers]{natbib}
\usepackage{doi}
\usepackage{hyperref}

\creflabelformat{*}{\textcolor{darkred}{#1}}
\title{How to Measure the Intelligence of Large Language Models?}

\author{{Nils K\"orber, Silvan Wehrli, Christopher Irrgang}\\
	Centre for Artificial Intelligence in Public Health Research \\
	Robert Koch Institute\\
	Berlin\\
	Germany\\
	\texttt{\{KoerberN, WehrliS, IrrgangC\}@rki.de} \\
}

\date{}
\renewcommand{\headeright}{}
\renewcommand{\undertitle}{}

\begin{document}

\twocolumn[
  \begin{@twocolumnfalse}
\maketitle

\vspace*{0.5cm}
  \end{@twocolumnfalse}
]

With the release of ChatGPT and other large language models (LLMs) the discussion about the intelligence, possibilities, and risks, of current and future models have seen large attention \cite{bengio2024managing, fei2022towards, schaeffer2024emergent}. 
This discussion included much debated scenarios about the imminent rise of so-called “super-human” AI \cite{mitchell2023we}, i.e., AI systems that are orders of magnitude smarter than humans. 
In the spirit of Alan Turing, there is no doubt that current state-of-the-art language models already pass his famous test \cite{jones2024people}. 
Moreover, current models outperform humans in several benchmark tests \cite{perrault2024artificial}, so that publicly available LLMs have already become versatile companions that connect everyday life, industry and science. 
Despite their impressive capabilities, LLMs sometimes fail completely at tasks that are thought to be trivial for humans. In other cases, the trustworthiness of LLMs becomes much more elusive and difficult to evaluate. Taking the example of academia, language models are capable of writing convincing research articles on a given topic with only little input.
Yet, the lack of trustworthiness in terms of factual consistency or the existence of persistent hallucinations in AI-generated text bodies has led to a range of restrictions for AI-based content in many scientific journals.
In view of these observations, the question arises as to whether the same metrics that apply to human intelligence can also be applied to computational methods and has been discussed extensively \cite{jones2024ai}. 
In fact, the choice of metrics has already been shown to dramatically influence assessments on potential intelligence emergence \cite{schaeffer2024emergent}. Here, we argue that the intelligence of LLMs should not only be assessed by task-specific statistical metrics, but separately in terms of qualitative and quantitative measures.

\section*{Quantitative and qualitative intelligence}

To evaluate the intelligence performance of LLMs, the problem can be easier accessed by separating the task into two separate metric classes: \textit{quantitative intelligence} and \textit{qualitative intelligence}. 
With \textit{quantitative intelligence} we refer to the model's inherent data storage and the model’s ability to navigate, use and remix this information, analogous to the knowledge of a human. In contrast, \textit{qualitative intelligence} refers to the ability to analyze, judge, and conclude from that data storage and novel information. At first glance, this broad separation may seem obvious, especially in the context of human intelligence. Nevertheless, there is still no standardized paradigm for evaluating intelligence for LLMs and similar AI systems. \par

\begin{figure*}[h]
\begin{center}
\includegraphics[width=\textwidth]{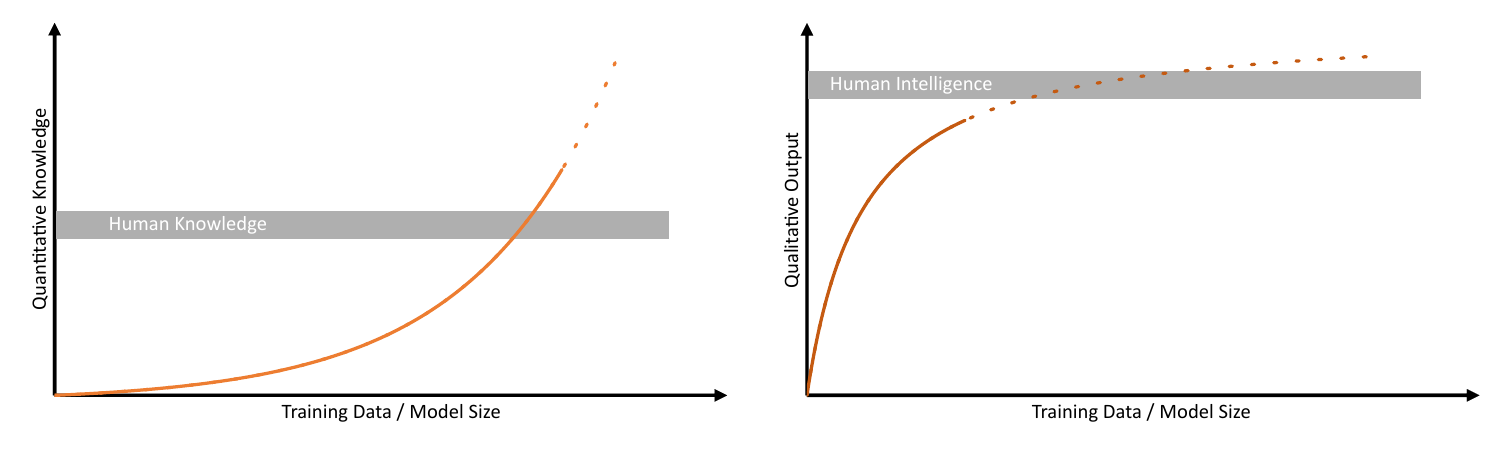}
\caption{Quantitative (left) and qualitative (right) performance of LLMs.}
\label{quanqual}
\end{center}

\end{figure*}

Quantitative intelligence includes the combined fields of expertise that are covered by an
LLM. As the current LLMs are trained on massive amounts of data,
which mostly come from internet sources, the quantitative capacities are enormous. A
reasonable conversation with an LLM is possible in dozens of languages and may
cover topics from knitting tips via ancient literature to quantum physics. At this stage, current
models likely already can fall back on more information than any human being (see Figure \ref{quanqual}).
To assess quantitative intelligence, a systematic test should be carried out that covers a wide
range of topics and different levels of detail within each topic. Quantitative measures are
directly dependent on the training data and describe the areas of specialization. Above that,
the quantitative capacity of a model provides better insight into whether the output of a model
is a simple rearrangement of training data or is actually concluded from context. In contrast to
humans, LLMs do not suffer from concentration loss and motivation and can be
evaluated quantitatively. Hence, instead of using a handful of questions as in a regular
examination test, a model might be analyzed on a larger scale with hundreds of thousands of
questions and tasks in a reasonably short time. Provided a comprehensive question catalog
covering the entire information spectrum from elementary school, common knowledge to
university curricula. The factual accuracy of the individual responses and the corresponding
metrics within each topic, across topics and over varying levels of complexity draw a picture
of the model's knowledge space. The massive multitask language understanding (MMLU) test covering almost 16,000 questions separated over 57 tasks is a step in the right direction. However, it does not explicitly differ between knowledge retrievel and problem solving capabilities \cite{hendrycks2020measuring}.
In order to examine the quantitative capabilities in detail, the questions should be chosen to focus exclusively information retrieval instead of reasoning, which should be addressed in a separate qualitative
test.\par

Current paradigms to evaluate human intelligence are based on the assumption that a subject
being able to complete a specific task is also able to complete similar tasks with lower or
comparable difficulty. If this generalizability holds true for LLMs is of ongoing
debate \cite{lu2021fantastically, yadlowsky2023pretraining} and could be represented as what we claim as qualitative intelligence. With the
term qualitative intelligence, we refer to capabilities such as reasoning, strategic actions and
drawing meaningful conclusions from data that were never seen during training. The last point
distinguishes whether a model only rearranges seen data or is truly able to solve unseen
problems, which is claimed to be an intelligence characteristic. Compared to quantitative
intelligence, qualitative measures require a whole different category of examination
techniques. However, given the partly closed source/weights of certain models and the sheer
amount of training data, that point is not easily addressable. A noteworthy example is the
current investigation on the persuasive capabilities of LLMs in staged debates.
By applying randomized control trials (RCTs), which are often used in clinical settings or causal
investigations, the machine versus human persuasiveness can be assessed in both qualitative
and quantitative measures \cite{salvi2024conversational}. The assessment of the performance in a crowdsource setting like the Chatbot Arena provides valuable tools for qualitative evaluation \cite{chiang2024chatbot}. Similar to standardised tools such as RCTs, a gold standard for qualitative assessments of intelligence is urgently needed but still missing.
Nevertheless, comparing the results in reasoning tests of
specialized models, such as the first transformer models with their successor to current
LLMs, the improvements in terms of qualitative measures are only gradual
\cite{nezhurina2024alice}. This is particularly true when you consider the model size and the training data,
which have increased by multiple orders of magnitude over the same time (Figure \ref{quanqual}). For a fair comparison of general reasoning capabilites between models, the quantitative aspect should be removed from the evaluation.\par

\section*{Computational growth versus intelligence growth}

The exponential growth of used training data with accompanying increase in model size will
likely put the models in the foreseeable future to a limit regarding data availability. As a thought
experiment, imagine training a model on the complete information on the internet, all text ever
written, and all thoughts every human ever said. How smart could such a model potentially
become given the current self-supervised training paradigm predicting a word in a
sentence? Quantitatively, the model would encompass the entire human knowledge.
Qualitatively it potentially could outsmart every person ever lived, being fed with countless
human ideas. However, we argue that the hypothetical qualitative improvement over humans
would still be on a comparable scale as the whole training basis inherently relies on human
thoughts and language from which the model can not escape. In the same way, a perfect
model for the language of an animal species, e.g., songbirds, would have difficulty explaining
Einstein's theory of relativity. The basis of large language models relies on their training data,
which is based on the mental capacity of the underlying species. While LLMs undoubtebly demonstrate impressive linguistic precision, it remains unclear how that translates into the broader cognitive capabilities often associated with language \cite{mahowald2024dissociating}. 
This raises the question of whether language alone is a sufficient means for acquiring such capabilities \cite{silver2021reward}. Furthermore, the debate persists as to how a simple reorganisation of huge amounts of data might not be sufficient to achieve results that resemble human behaviour\cite{mitchell2023debate}.

\par
Relating these thoughts to the notion of “super-human” AI, we see that current and future
state-of-the-art LLMs can exceed human intelligence quantitatively with ease.
This is especially the case, when AI tasks are combined with their usually very short inference
time. On the contrary, future LLMs that have a qualitative improvement orders of
magnitudes higher than humans seem still very unlikely or even impossible with current learning
paradigms. On top, truly emergent intelligence properties of such models might even be
invisible to current measures. To ultimately prove or disprove emergent intelligence in
concrete settings or for specific LLMs, nuanced and combined attention to both quantitative and qualitative intelligence
measures will be crucial. The development of frameworks that address these issues should
receive dedicated awareness by the research communities.\par
Undoubtedly, even without singular intelligence growth of LLMs, their societal
impact is already tremendous \cite{perrault2024artificial}. With growing quantitative intelligence, models can be
applied to a plethora of domains combining expertise from various fields. Tasks that currently
require a team with different backgrounds could be handled by a single LLM in
the future. Thus, a simple prompt can be sufficient to launch a new campaign, create a movie,
or plan a medical study. These advancements are fundamental for accelerating existing
processes. Access to superior technology or new physical concepts, on the other hand, is not
on the immediate horizon. Accordingly, the creation of an uncontrollable superintelligence may not be the most imminent threat, but large-scale job losses, misinformation and election interference are. \par
In summary, current and future developments in LLMs will lead to models that
represent a significant part of the entire human knowledge, but which might not quickly
surpass the qualitative output of humans. To address this bivalence and to reliably examine
emergent intelligence properties or even “super-human” behavior, the intelligence
performance of models should be accessed with separated quantitative and qualitative
intelligence metrics.

\bibliographystyle{unsrt}  
\bibliography{ref}

\end{document}